\PassOptionsToPackage{table}{xcolor}
\documentclass[10pt,twocolumn,letterpaper]{article}

\usepackage[pagenumbers]{cvpr} 

\usepackage[dvipsnames]{xcolor}

\definecolor{cvprblue}{rgb}{0.21,0.49,0.74}
\usepackage[pagebackref,breaklinks,colorlinks,citecolor=cvprblue]{hyperref}
\usepackage{multirow}
\usepackage{graphicx}
\usepackage{caption}
\usepackage{subcaption}
\usepackage{verbatim}
\usepackage{arydshln}
\usepackage{array}

\def\eg{{\em e.g.}}
\def\ie{{\em i.e.}}

\definecolor{mygray}{gray}{.9}
\newcommand{\improve}[1]{\scriptsize{\textcolor[RGB]{61,145,64}{~(#1)}}}

\title{Context-Guided Spatio-Temporal Video Grounding}

\author{
    Xin Gu$^{1,3*}$, 
    Heng Fan$^{2*}$,
     Yan Huang$^2$,
    Tiejian Luo$^1$,
    Libo Zhang$^{1,3\dagger}$\\
    $^1$University of Chinese Academy of Sciences, Beijing, China\\
    $^2$Department of Computer Science and Engineering, University of North Texas, Denton, USA\\
    $^3$Institute of Software, Chinese Academy of Sciences, Beijing, China
}

\begin{document}
	
\maketitle

\def\thefootnote{\textdagger}\footnotetext{Corresponding author: Libo Zhang (libo@iscas.ac.cn).}
\def\thefootnote{\arabic{footnote}}

\begin{abstract}
	
	Spatio-temporal video grounding (or STVG) task aims at locating a spatio-temporal tube for a specific instance given a text query. Despite advancements, current methods easily suffer the distractors or heavy object appearance variations in videos due to insufficient object information from the text, leading to degradation. Addressing this, we propose a novel framework, context-guided STVG (CG-STVG), which mines discriminative instance context for object in videos and applies it as a supplementary guidance for target localization. The key of CG-STVG lies in two specially designed modules, including instance context generation (ICG), which focuses on discovering visual context information (in both appearance and motion) of the instance, and instance context refinement (ICR), which aims to improve the instance context from ICG by eliminating irrelevant or even harmful information from the context. During grounding, ICG, together with ICR, are deployed at each decoding stage of a Transformer architecture for instance context learning. Particularly, instance context learned from one decoding stage is fed to the next stage, and leveraged as a guidance containing rich and discriminative object feature to enhance the target-awareness in decoding feature, which conversely benefits generating better new instance context for improving localization finally. Compared to existing methods, CG-STVG enjoys object information in text query and guidance from mined instance visual context for more accurate target localization. In our experiments on three benchmarks, including HCSTVG-v1/-v2 and VidSTG, CG-STVG sets new state-of-the-arts in m\_tIoU and m\_vIoU on all of them, showing its efficacy. The code will be released at \url{https://github.com/HengLan/CGSTVG}.
	
\end{abstract}

\section{Introduction}
\label{sec:intro}

Spatio-temporal video grounding task, or \emph{\textbf{STVG}}, is recently introduced in~\cite{STGRN} and aims to localize the object of interest in an untrimmed video with a spatio-temporal tube (formed by a sequence of bounding boxes) given a \emph{free-form} textual query. It is a challenging multimodal task which is involved with learning and understanding spatio-temporal visual representations in videos and their connections to the linguistic representation of text. Due to the importance in multimodal video understanding, STVG has drawn increasing attention in recent years (\eg,~\cite{STGRN,OAMBRN,hcstvg,STVGBert,TubeDETR,STCAT,csdvl}).

\begin{figure*}[t]
	\centering
	\includegraphics[width=0.9\linewidth]{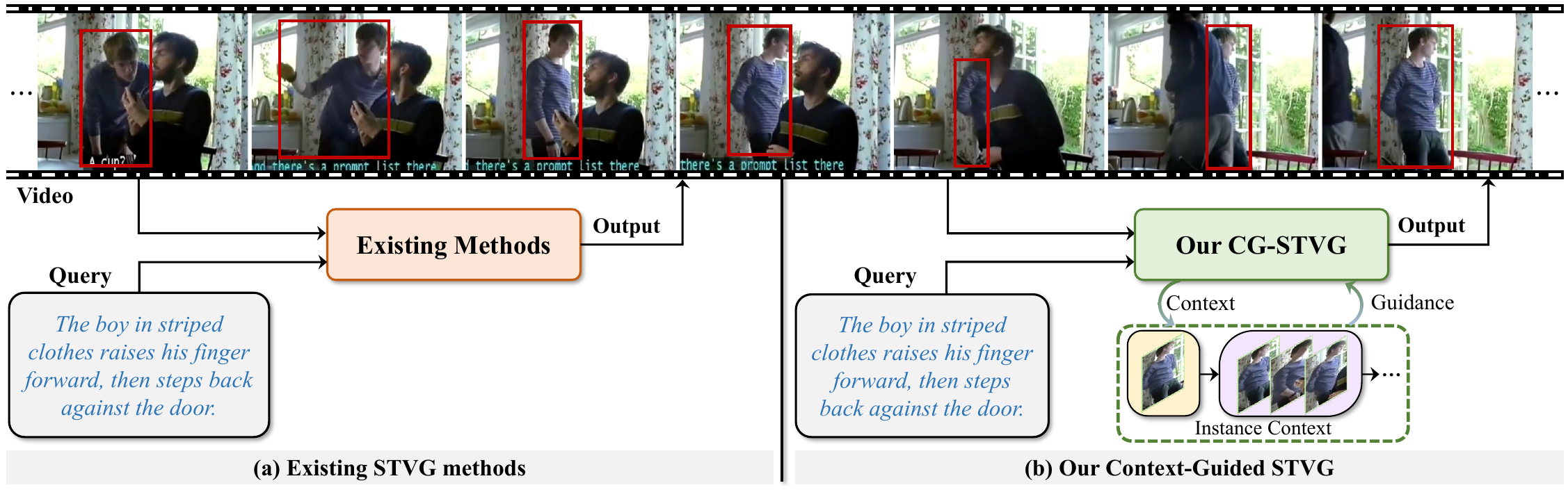}
	\caption{Comparison between (a) existing methods that localize the target using object information from text query and (b) our CG-STVG that enjoys object information from text query and guidance from mined instance context for STVG. \emph{Best viewed in color for all figures}.}
	\label{fig:comp}
 \vspace{-2mm}
\end{figure*}

Current methods usually use the given textual expression as the \emph{only} cue for retrieving object in videos (see Fig.~\ref{fig:comp} (a)). Despite progress, they may degrade in complex scenes (\eg, in presence of distractors, or severe appearance changes, or both in videos), because text query is \emph{insufficient} to describe and distinguish the foreground object in these cases. To alleviate this problem, one straightforward solution is to enhance the textual query by including more fine-grained linguistic description. However, there may exist several issues. \emph{First}, this needs reconstruction of text queries for all objects with longer detailed descriptions, which is laborious as well as expensive. \emph{Second}, longer text query will result in more computational overheads for training and inference. \emph{Third}, although the text query can be enhanced with more details, it might still be hard to comprehensively describe certain visual details~\cite{zhou2022conditional}. Thus, it is natural to ask: \emph{Is there any other way, besides enhancing text query, that improves efficiently, effectively, and friendly spatio-temporal video ground?}

We answer \emph{yes}! Instead of enhancing the text query, we propose to exploit \emph{visual information} of the object to offer a guidance, directly from the \emph{vision perspective}, for improving STVG. As indicated in the famous saying, ``\emph{A Picture Is Worth a Thousand Words}'', visual cues can provide richer information with description granularity about the target object. Nevertheless, for the STVG task, there is \emph{no} additional \emph{external} visual information allowed, besides the text query, for target localization. So, \emph{where to acquire the desired visual information for improving STVG?} 

\begin{figure*}[t]
	\centering
        \includegraphics[trim=5 112 0 90, clip, width=0.94\linewidth]{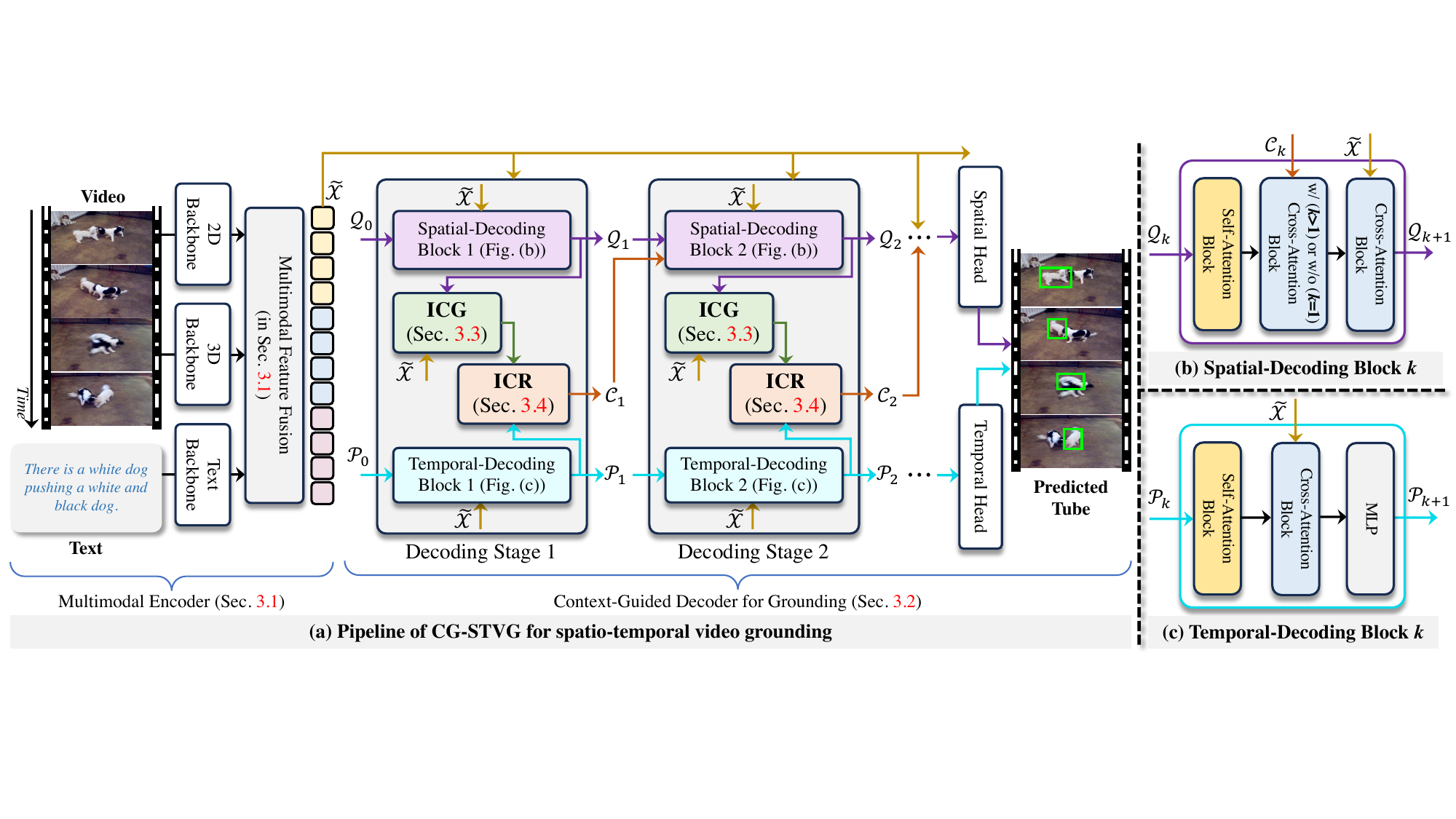}
	\caption{Overview of our method, which consists of a multimodal encoder for feature extraction and a context-guided decoder by cascading a set of decoding stages for grounding. In each decoding stage, instance context is mined to guide query learning for better localization.}
	\label{fig:framework}
	\vspace{-3mm}
\end{figure*}

\emph{From the video itself!} In this paper, we introduce a novel framework, context-guided STVG or CG-STVG, that mines \emph{internally} discriminative visual context information from a video for the object, and uses it as a supplementary guidance to improve target localization (see Fig.~\ref{fig:comp} (b)). The crux of CG-STVG lies in two crucial modules, including instance context generation (or ICG) and instance context refinement (or ICR). ICG focuses on discovering visual information of the object. Specifically, ICG first estimates potential regions for the foreground and then uses them to extract contextual information of both appearance and motion from the visual features. Considering there might exist noises in contextual features that are irrelevant or even harmful for the localization due to inaccurate foreground region estimation, ICR is leveraged to eliminate the useless information. Concretely, it adopts a joint temporal-spatio filtering way based on the temporal and spatio relevance scores to suppress irrelevant features, greatly enhancing the context for localization. In this work, we adopt DETR-similar architecture~\cite{carion2020end} to implement CG-STVG. During video grounding, ICG, together with the ICR, are deployed at each of the decoding stage for instance context learning. Particularly, the instance context learned from one decoding stage is fed to the next stage, and used as a supplementary guidance containing rich and discriminative object information to enhance target-awareness of decoding feature, which in turn benefits generating better new instance context for improving the localization finally. Fig.~\ref{fig:framework} illustrates the architecture of CG-STVG.

To our best knowledge, CG-STVG is the first to mine instance visual context from the videos to guide STVG. Compared with existing approaches, CG-STVG can leverage the object information from both text query, as in current methods, and guidance from its mined instance context for more accurate target localization. To validate its effectiveness, we conduct extensive experiments on three datasets, including HCSTVG-v1/-v2~\cite{hcstvg} and VidSTG~\cite{wheredoseit}, CG-STVG outperforms existing methods and sets new state-of-the-arts in m\_tIoU and m\_vIoU on all of these benchmarks, evidencing the efficacy of guidance from instance context for STVG.

In summary, the main contributions are as follows:
\vspace{0.1em}
\begin{itemize}
	\setlength{\itemsep}{2pt}
	\setlength{\parsep}{2pt}
	\setlength{\parskip}{2pt}
	
	\item[$\spadesuit$] \emph{We introduce CG-STVG, a novel and simple approach for improving STVG via mining instance visual context from the video to guide target localization. 
	}
	
	\item[$\heartsuit$] \emph{We propose an instance context generation module (ICG) to discover visual context information of the object.}
	
	\item[$\clubsuit$] \emph{An instance context refinement (ICR) module is presented to improve the context of object by eliminating irrelevant contextual features, greatly enhancing the performance.}
	
	\item[$\diamondsuit$] \emph{In extensive experiments on three benchmarks, including HCSTVG-v1/-v2~\cite{hcstvg} and VidSTG~\cite{wheredoseit}, CG-STVG sets new state-of-the-arts, showing the effectiveness.}
\end{itemize}

\section{Related Work}
\label{sec:related_work}

\noindent
\textbf{Spatio-temporal video grounding.} Spatio-temporal video grounding~\cite{hcstvg} aims to generate a spatio-temporal tube for a target given its text query.
Early methods (\eg,~\cite{STGRN, OAMBRN, hcstvg}) mainly follow a two-stage paradigm, which leverages a pretrained detector to obtain the candidate region proposals and then finds the correct region proposals through the designed network. The main issue of these methods is the heavy reliance on pre-trained detectors, and the performance is restricted by a detector's own limitations. Differently, recent works (\eg,~\cite{STVGBert, TubeDETR,STCAT,csdvl}) adopt a one-stage paradigm, directly generating spatio-temporal object proposals without relying on any pre-trained object detectors. The method of~\cite{STVGBert} is the first of this kind, which leverages the visual-linguistic transformer to generate a spatio-temporal object tube corresponding to the textual sentence. Inspired by the success of the model for text-conditioned object detection~\cite{mdetr}, the method in~\cite{TubeDETR} introduces a spatio-temporal transformer decoder together with a video-text encoder for STVG. The approach of~\cite{STCAT} utilizes a multi-modal template as the global objective to deal with the inconsistency issue for improvement. The work of~\cite{csdvl} proposes to explore static appearance and dynamic motion cues collaboratively for target localization, showing promising results.

In this paper, we focus on exploiting discriminative instance visual context from the video and adopt it as a guidance for target localization. \textbf{\emph{Different}} from existing STVG approaches (\eg,~\cite{csdvl, STVGBert,STCAT, TubeDETR}) which explore object information only from the text query for localization, the proposed CG-STVG is able to leverage both textual cue and object guidance from the mined instance context, significantly enhancing the STVG performance and outperforming other methods, particularly in complicated scenarios with similar distractors or large appearance changes.

\vspace{0.2em}
\noindent
\textbf{Temporal Grounding.}
Temporal grounding aims at locating and understanding specific objects or events in a video. In specific, the goal is to jointly comprehend visual information and natural language descriptions in a video, identify the target objects, understand their actions and environments, and link them with the natural language descriptions. Relevant to but different than the STVG, temporal grounding does not require bounding box localization of the target. Numerous approaches (\eg,~\cite{locvtp,drft,mun2020local,hao2022can,wang2023protege,zhang2023text,lmmg}) have been introduced recently. For example, the algorithm of~\cite{lmmg} proposes an effective strategy to avoid the long-form burden by applying a guidance model for grounding time. The approach of~\cite{locvtp} leverages cross-modal contrastive learning at coarse-grained (video-sentence) and fine-grained (clip-word) levels for grounding. The work in~\cite{drft} designs a multimodal framework to learn complementary features from images, flow, and depth for the temporal grounding. \textbf{\emph{Different}} than these methods, we focus on the more challenging STVG that spatially and temporally localizes the object.

\vspace{0.2em}
\noindent
\textbf{Vision-Language Modeling.}
Vision-language modeling is to simultaneously process visual and linguistic information for joint multimodal understanding and it has drawn extensive interest from researchers in various tasks such as visual question answering~\cite{antol2015vqa,jiang2020defense,yu2019deep,chen2022rethinking,le2020hierarchical,shao2023prompting}, image captioning~\cite{you2016image,aneja2018convolutional,huang2019attention}, video captioning~\cite{zhou2018end,iashin2020multi,shen2023accurate,seo2022end,yang2023vid2seq}, text-to-image generation~\cite{li2019controllable,ramesh2021zero}, visual-language tracking~\cite{guo2022divert,zhou2023joint}, etc. \emph{\textbf{Different}} from the above tasks, we focus on modeling vision and language for spatio-temporal target localization.

\section{The Proposed Method}

\textbf{Overview.} In this work, we introduce CG-STVG by mining the discriminative visual context of object and using it as a guidance to improve the target localization. Inspired by the DETR~\cite{carion2020end} for detection, CG-STVG employs an encoder-decoder architecture, which comprises the multimodal encoder (Sec.~\ref{mmef}) and the context-guided decoder (Sec.~\ref{grounding}). As illustrated in Fig.~\ref{fig:framework}, the encoder aims at generating multimodal visual-linguistic feature that contains object information from text query, which is sent to the context-guided decoder for target localization guided by instance context learned with ICG (Sec.~\ref{icg}) and ICR (Sec.~\ref{icr}).

\subsection{Multimodal Encoder}
\label{mmef}

The multimodal encoder is to generate a robust multimodal feature for the target localization in decoder, and consists of visual and textual feature extraction and fusion as follows.

\vspace{0.2em}
\noindent 
\textbf{Visual Feature Extraction.} To leverage rich cues from the videos, we extract both the appearance and motion features. In specific, we first sample a set of frames $\mathcal{F} = \{f_i\}_{i=1}^{N_{v}}$ of length $N_{v}$ from the video, and then utilize ResNet-101~\cite{resnet} for appearance feature extraction and VidSwin~\cite{vidswin} for motion feature extraction, respectively. We denote the appearance feature as $\mathcal{V}_{a}=\{v^{a}_i\}_{i=1}^{N_{v}}$, where $v_i^{a} \in \mathbb{R}^{H \times W \times C_a}$ with $H$, $W$, and $C_a$ the height, width and channel dimensions. Similarly, we denote the motion feature as $\mathcal{V}_{m}=\{v^{m}_i\}_{i=1}^{N_{v}}$, where $v_i^{m} \in \mathbb{R}^{H \times W \times C_m}$ with $C_m$ the channel dimension.

\vspace{0.2em}
\noindent 
\textbf{Textual Feature Extraction.} We adopt RoBERTa~\cite{roberta} for textual feature extraction. We first tokenize query to obtain a word sequence $\mathcal{W} = \{w_i\}_{i=1}^{i=N_t}$ and then apply RoBERTa to produce an embedding sequence $\mathcal{T} = \{t_i\}_{i=1}^{i=N_t}$, where $t_i \in \mathbb{R}^{C_t}$ with $C_t$ the word embedding dimension.

\vspace{0.2em}
\noindent 
\textbf{Multimodal Feature Fusion.} STVG is a multimodal task. To enhance feature representation, we perform multimodal fusion of the appearance feature $\mathcal{V}_{a}$, motion feature $\mathcal{V}_{m}$, and text feature $\mathcal{T}$. Specifically, we first map $\mathcal{V}_{a}$, $\mathcal{V}_{m}$ and   $\mathcal{T}$ to the same channel number through linear projection and then concatenate corresponding features to obtain the representation of multimodal features $\mathcal{X}=\{x_i\}_{i=1}^{N_v}$ as follows,
\begin{equation*}
	x_i = [\underbrace{v^a_{i_1}, v^a_{i_2}, ..., v^a_{i_{H \times W}}}_{\text{appearance features $v^a_i$}},\underbrace{v^m_{i_1}, v^m_{i_2}, ..., v^m_{i_{H \times W}}}_{\text{motion features $v^m_i$}}, \underbrace{t_{1}, t_{2}, ..., t_{N_t}}_{\text{textual features $\mathcal{T}$}}]
\end{equation*}
where $x_i$ is the multimodal feature in frame $i$. Then, we add position embedding $\mathcal{E}_{pos}$ and type embedding $\mathcal{E}_{typ}$ to $\mathcal{X}$ by
\begin{equation*}
	\mathcal{X}'=\mathcal{X} + \mathcal{E}_{pos} + \mathcal{E}_{typ}
\end{equation*}
Finally, we perform multimodal feature fusion by applying a self-attention encoder on $\mathcal{X}'$ as follows,
\begin{equation*}
	\mathcal{\tilde{X}} = \mathtt{SAEncoder}(\mathcal{X}')
\end{equation*}
where $\mathcal{\tilde{X}}$ is the enhanced multimodal feature for decoding, and $\mathtt{SAEncoder}(\cdot)$ the self-attention encoder with $L$ ($L$=6) standard self-attention encoder blocks~\cite{vaswani2017attention}. Please refer to \textbf{supplementary material} for architecture of $\mathtt{SAEncoder}(\cdot)$.

\subsection{Context-Guided Decoder for Grounding}
\label{grounding}

CG-STVG designs a context-guided decoder with $K$ stages in a cascade for grounding as in Fig.~\ref{fig:framework} (a). Since CG-STVG needs to locate target spatially and temporally, each decoding stage has two blocks, including a \emph{spatial-decoding block} (SDB) and a \emph{temporal-decoding block} (TDB), for spatial and temporal feature learning. In each stage (except for the first), instance context by ICG and ICR (see later) is applied as a guidance with rich visual cue to enhance the query feature, which is in turn used to generate new instance context.

Specifically, let $\mathcal{Q}_{k-1}$=${\{q_i^{k-1}\}_{i=1}^{N_v}}$ denote spatial query features for $N_v$ frames and $\mathcal{P}_{k-1}$=${\{p_i^{k-1}\}_{i=1}^{N_v}}$the  temporal query features sent to the $k^{\text{th}}$ ($1<k\leq K$) decoding stage. $\mathcal{Q}_0$ and $\mathcal{P}_0$ fed to the first decoding stage are initialized following DETR~\cite{carion2020end}. Then, in decoding stage $k$, we use SDB$_{k}$ to learn query feature $\mathcal{Q}_{k}$ using instance context $\mathcal{C}_{k-1}$ from  decoding stage ($k$-1) as a guidance and multimodal feature $\mathcal{\tilde{X}}$ from the encoder. As in Fig.~\ref{fig:framework} (b), SDB$_{k}$ contains three components with one self-attention and two cross-attention blocks. The self-attention block is to enhance query features by interacting them. The former cross-attention block aims to guide query features using $\mathcal{C}_{k-1}$, while the later is for learning object position information from $\mathcal{\tilde{X}}$. The process of SDB$_{k}$ for learning $\mathcal{Q}_k$ can be formulated as follows,
\begin{equation*}
	\begin{split}
		\mathcal{Q}_k &= \mathtt{SDB}_{k}(\mathcal{Q}_{k-1}, \mathcal{C}_{k-1}, \mathcal{\tilde{X}}) \\ &=\mathtt{CA}(\mathtt{CA}(\mathtt{SA}(\mathcal{Q}_{k-1}), \mathcal{C}_{k-1}), \mathcal{\tilde{X}})
	\end{split}
\end{equation*}
where $\mathtt{SA}(\textbf{z})$ denotes the self-attention block with $\textbf{z}$ generating query/key/value, and $\mathtt{CA}(\textbf{z}, \textbf{u})$ the cross-attention block with $\textbf{z}$ generating query and $\textbf{u}$ key/value, as in~\cite{vaswani2017attention}. Due to limited space, please see \textbf{supplementary material} for detailed architectures. For SDB$_{1}$, because the instance context does not exist, $\mathcal{Q}_1$ is learned as follows,
\begin{equation*}
		\mathcal{Q}_1 = \mathtt{SDB}_{1}(\mathcal{Q}_{0}, \mathcal{\tilde{X}}) = \mathtt{CA}(\mathtt{SA}(\mathcal{Q}_{0}), \mathcal{\tilde{X}})
\end{equation*}

\begin{figure}[t]
	\centering
	\includegraphics[width=0.92\linewidth]{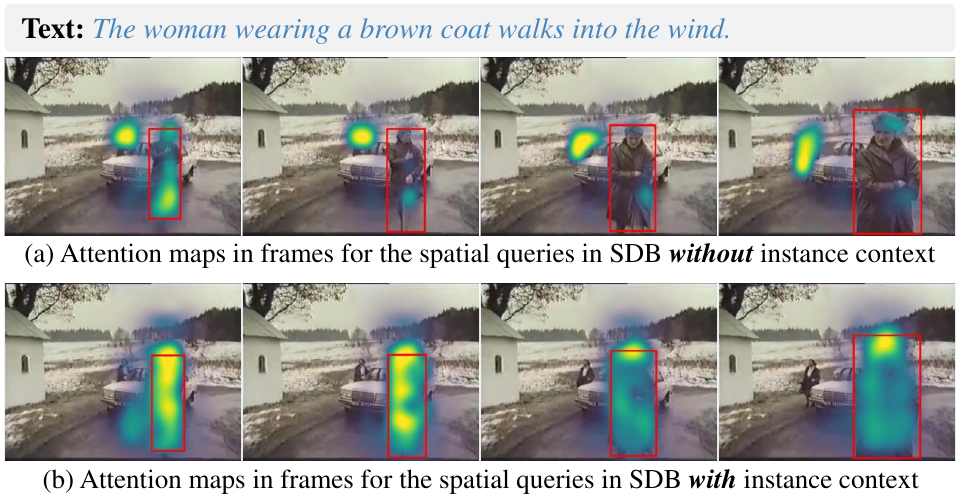}
	\caption{Attention maps for spatial queries in video frames in the spatial-decoding block \emph{without} (image (a)) and \emph{with} our proposed instance context (image (b)). We can clearly see that our instance context effectively improves target-awareness in the spatial queries and thus the target position information learning for localization. The red boxes indicate the foreground object to localize.}
	\label{fig:att}
	\vspace{-3mm}
\end{figure}

In decoding, the spatial query feature aims to learn object information progressively from $\mathcal{\tilde{X}}$. In our SDB, the spatial query feature is guided by the visual context of the object to enhance its \emph{target-awareness} in \emph{vision perspective} such that it can explicitly exploit rich and discriminative visual cues to learn more accurate position information from $\mathcal{\tilde{X}}$ for better target localization, even when text cannot well describe the object, which significantly differs than existing methods (\eg,~\cite{STVGBert, TubeDETR,STCAT,csdvl}).

Similarly in decoding stage $k$, the temporal query feature $\mathcal{P}_{k}$ is learned by TDB$_{k}$ which consists of self-attention and cross-attention blocks followed by the MLP, as in Fig.~\ref{fig:framework} (c). The process for learning $\mathcal{P}_{k}$ can be expressed as follows, 
\begin{equation*}
	\mathcal{P}_k = \mathtt{TDB}_{k}(\mathcal{P}_{k-1}, \mathcal{\tilde{X}}) = \mathtt{MLP}(\mathtt{CA}(\mathtt{SA}(\mathcal{P}_{k-1}), \mathcal{\tilde{X}}))
\end{equation*}
Notice that, instance context $\mathcal{C}_{k-1}$ is not used in TDB, as it mainly works to localize target when it exists in the frames, instead of detecting if the object exists or not. When applying instance context in TDB, it even cause slight performance drop. Thus, instance context is only applied in STB.

Once generating $\mathcal{Q}_{k}$ and $\mathcal{P}_{k}$, they are used to learn new instance context $\mathcal{C}_{k}$ in decoding stage $k$ with already acquired object position and frame information using ICG and ICR (as explained later), which will be applied to guide further query learning in subsequent stages for improving target-awareness and position information learning, as evidenced in Fig.~\ref{fig:att}, in a progressive way. In the decoding stage $K$, the learned $\mathcal{Q}_{K}$ and $\mathcal{P}_{K}$ are fed to two heads to predict the final object boxes $\mathcal{B}_K=\{b_i\}_{i=1}^{N_v}$, where $b_i\in R^{4}$ denotes the central position, width and height of the prediction box, and the start and end probabilities of each frame $\mathcal{H}_K =\{(h_i^s, h_i^e)\}_{i=1}^{N_v}$, where the start and end times are determined by the maximum joint start and end probability.

\subsection{Instance Context Generation (ICG)}
\label{icg}

\begin{figure}[t]
	\centering
	\includegraphics[width=\linewidth]{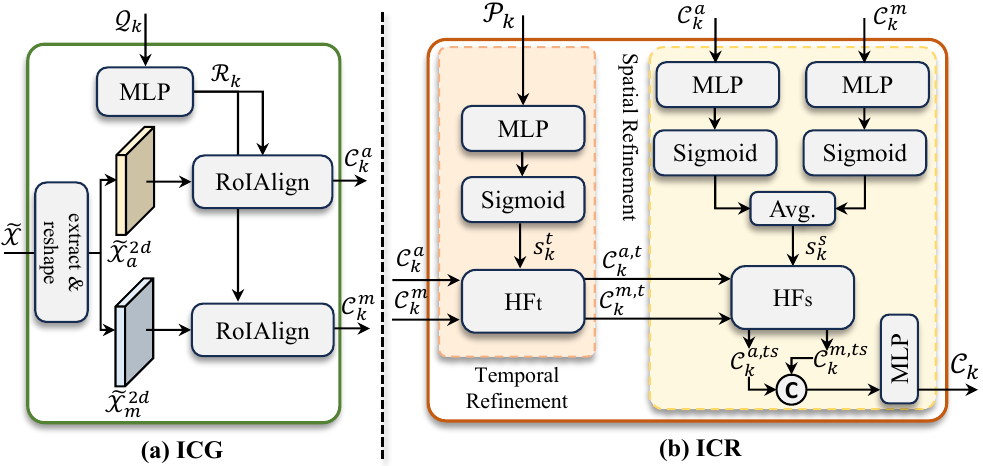}
	\caption{Illustration ICG (image (a)) and ICR (image (b)).}
	\label{fig:icg}
	\vspace{-3mm}
\end{figure}

To exploit instance context in the video, we introduce a simple yet effective module, termed instance context generation (ICG). Specifically, ICG is deployed in each decoding stage $k$ of the context-guided decoder and takes the spatial query feature $\mathcal{Q}_{k}$ learned from SDB$_{k}$ to discover the potential features of the foreground (\ie, the so-called \emph{instance context}), as in Fig.~\ref{fig:icg} (a). The \textbf{intuition} is that, during the progressive video grounding for localization, $\mathcal{Q}_{k}$ has gradually learned more and more position information for the object and thus can be employed to find target regions in a video, which are used to further mine object features in the decoding stage $k$. 

To this end, we first utilize a three-layer MLP in ICG to transform $\mathcal{Q}_{k}$ into foreground regions $\mathcal{R}_{k}$ as follows,
\begin{equation}\label{mlp}
	\mathcal{R}_{k} = \mathtt{MLP}(\mathcal{Q}_{k}) = \{r^k_i\}_{i=1}^{N_v}
\end{equation}
where $r^k_i\in R^{4}$ denotes estimated object center and scale in frame $i$. Then with $\mathcal{R}_{k}$, we leverage it to extract the corresponding foreground features, including both appearance and motion features. In specific,  we first extract the appearance and motion features, denoted by $\mathcal{\tilde{X}}_{a}$ and $\mathcal{\tilde{X}}_{m}$, from the multimodal feature $\mathcal{\tilde{X}}$, and then reshape them into 2D feature maps $\mathcal{\tilde{X}}_{a}^{2d}$=$\text{reshape}(\mathcal{\tilde{X}}_{a})$ and $\mathcal{\tilde{X}}_{m}^{2d}$=$\text{reshape}(\mathcal{\tilde{X}}_{m})$. After that, we use RoIAlign~\cite{he2017mask} to extract appearance and motion instance context as follows,
\begin{equation*}
	\begin{split}
		\mathcal{C}_{k}^{a} = \mathtt{RoIAlign}(\mathcal{\tilde{X}}_{a}^{2d}, \mathcal{R}_{k}) \;\;\;
		\mathcal{C}_{k}^{m} = \mathtt{RoIAlign}(\mathcal{\tilde{X}}_{m}^{2d}, \mathcal{R}_{k})
	\end{split}
\end{equation*}
where $\mathcal{C}_{k}^{a}$ denotes the appearance instance context and $\mathcal{C}_{k}^{m}$ the motion instance context. $\mathcal{C}_{k}^{a}$ mainly encompass various rich visual attributes of the target, such as shape, texture and color, while $\mathcal{C}_{k}^{a}$ predominantly captures motion properties of the object, including speed and trajectory. Both of these two context are beneficial to enhance the target-awareness, enhancing target-awareness in spatial query feature for better target position learning.

\begin{figure}[t]
	\centering
        \includegraphics[trim=0 300 390 0, clip, width=1.0\linewidth]{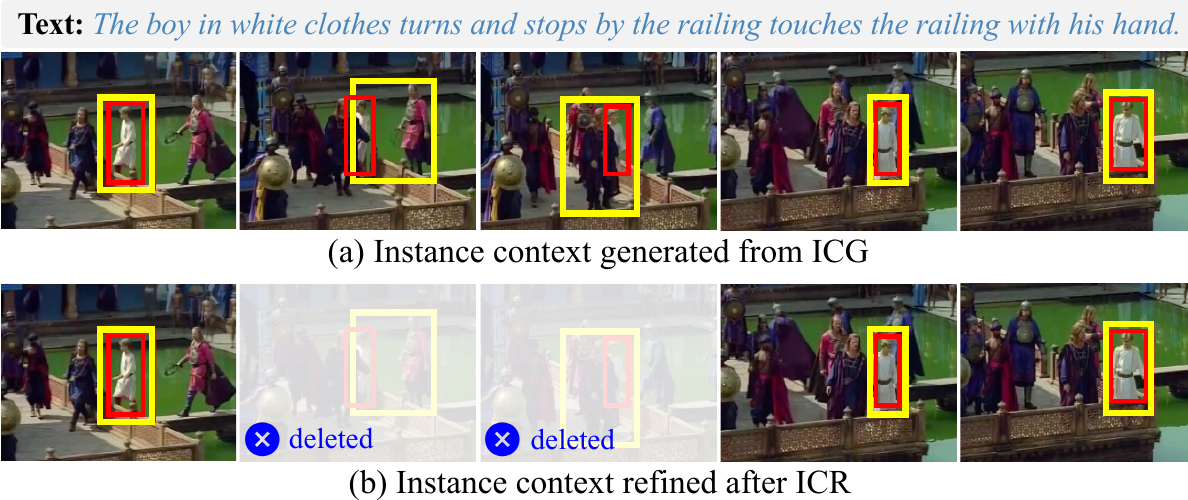}
	\caption{Illustration of ICR for context refinement. The red boxes indicate the foreground, while yellow boxes the instance context. We can see that, our ICR is able to help eliminate irrelevant features in the initial instance context generated from ICG.}
	\label{fig:contref}
	\vspace{-3mm}
\end{figure}

\subsection{Instance Context Refinement (ICR)}
\label{icr}

Considering that the estimated foreground regions may contain noise because the target position information in $\mathcal{Q}_{k}$ is not enough, the instance visual context of $\mathcal{C}_{k}^{a}$ and $\mathcal{C}_{k}^{m}$ might contain irrelevant and even harmful features, and thus is degraded. To remedy, we further present the instance context refinement module (ICR) to refine $\mathcal{C}_{k}^{a}$ and $\mathcal{C}_{k}^{m}$ for better final instance context by eliminating irrelevant features. In particular, we introduce a two-level temporal-spatial joint refinement mechanism in ICR. In the first level, instance context is refined by a temporal filter with temporal-confidence of each feature. Then, at the second level, a spatial filter is designed to suppress irrelevant features. Fig.~\ref{fig:icg} (b) shows the architecture of ICR, which is detailed as follows.

\textbf{Temporal Refinement.} Because instance context is only related to the object in a certain temporal window, instead of the whole video, we leverage the temporal query feature $\mathcal{P}_{k}$ in decoding stage $k$ to calculate the confidence score of each frame being relevant to the object. Specifically, we simply apply an MLP module followed by a Sigmoid function to transform $\mathcal{P}_{k}$ to the temporal confidence scores as follows,
\begin{equation*}
	s_{k}^{t}=\mathtt{Sigmoid}(\mathtt{MLP}(\mathcal{P}_{k}))
\end{equation*}
where $s_{k}^{t}\in R^{N_v}$ represents the temporal confidence scores. The higher the $s_{k}^{t}(i)$ is, the more relevant the instance feature in frame $i$ is. To eliminate irrelevant feature, we design a filter to drop instance context features with temporal confidence scores lower than a preset threshold $\theta^{t}$ as follows,
\begin{equation*}
	\mathcal{C}_{k}^{a,t}=\mathtt{HF_{t}}(\mathcal{C}_{k}^{a}, s_{k}^{t}, \theta^{t}) \;\;\;\;\;\; \mathcal{C}_{k}^{m,t}=\mathtt{HF_{t}}(\mathcal{C}_{k}^{m}, s_{k}^{t}, \theta^{t})
\end{equation*}
where $\mathcal{C}_{k}^{a,t}$ and $\mathcal{C}_{k}^{m,t}$ are refined instance context. $\mathtt{HF}_{t}$ is a high-pass filter that passes instance context features of $\mathcal{C}_{k}^{a}$ and $\mathcal{C}_{k}^{m}$ with confidence scores greater than $\theta^{t}$.

\textbf{Spatial Refinement.} Different from the temporal refinement, spatial refinement aims to measure the quality of context features $\mathcal{C}_{k}^{a}$ and $\mathcal{C}_{k}^{m}$ from spatial dimension. To this end, we apply two MLP modules with each followed by a Sigmoid function to compute the spatial appearance and motion confidence scores, which are averaged to obtain the final spatial confidence scores, as follows,
\begin{equation*}
	s_{k}^{s}=(\mathtt{Sigmoid}(\mathtt{MLP}(\mathcal{R}_{k}^{a}))+\mathtt{Sigmoid}(\mathtt{MLP}(\mathcal{R}_{k}^{m})))/2
\end{equation*}
where $s_{k}^{s}\in R^{N_v}$ represents spatial confidence scores which are measured using the predicted IoU confidence~\cite{jiang2018acquisition}, originally used for detection. To suppress irrelevant features, we drop features in $\mathcal{C}_{k}^{a,t}$ and $\mathcal{C}_{k}^{m,t}$ with temporal confidence scores lower than a preset threshold $\theta_{s}$ as follows,
\begin{equation*}
	\mathcal{C}_{k}^{a,ts}=\mathtt{HF_{s}}(\mathcal{C}_{k}^{a,t}, s_{k}^{t}, \theta^{s}) \;\;\;\;\; \mathcal{C}_{k}^{m,ts}=\mathtt{HF_{s}}(\mathcal{C}_{k}^{m,t}, s_{k}^{t}, \theta^{s})
\end{equation*}
where $\mathcal{C}_{k}^{a,ts}$ and $\mathcal{C}_{k}^{m,ts}$ are refined instance context. $\mathtt{HF}_{s}$ is a high-pass filter that passes instance context features with confidence scores greater than $\theta^{s}$. Fig.~\ref{fig:contref} illustrates the instance refinement by ICR.

\textbf{Final Instance Context.} After the two-level refinement, we concatenate $\mathcal{C}_{k}^{a,ts}$ and $\mathcal{C}_{k}^{m,ts}$ and apply an MLP module to obtain the final instance context $\mathcal{C}_{k}$ as in Fig.~\ref{fig:icg}, which is used to improve target position learning in the next stage.

\subsection{Optimization}

Given a video and its text, after the $k^{\text{th}}$ decoding stage, we predict: (1) start timestamps $\mathcal{H}_k^s = \{h_i^s\}_{i=1}^{N_v}$ and end timestamps $\mathcal{H}_k^e = \{h_i^e\}_{i=1}^{N_v}$ of the video clip related to text, (2) bounding box $\mathcal{B}_k=\{b_i\}_{i=1}^{N_v}$ of the object on which the text focuses, (3) temporal and spatial confidence scores $s_k^t$ and $s_k^s$ in context refinement. In training, given groundtruth start timestamps $\mathcal{H}_s^*$, the end timestamps $\mathcal{H}_e^*$, the bounding box sequence $\mathcal{B}^*$. 

For temporal grounding, the KL divergence and binary cross-entropy are used as the loss function and the losses of start and end times are computed as follows,
\begin{equation*}
\begin{split}
    \mathcal{L}_k^{t} = & \lambda_s \mathcal{L}_{KL}(\mathcal{H}_s^*,\mathcal{H}_k^s) + \lambda_e \mathcal{L}_{KL}(\mathcal{H}_e^*,\mathcal{H}_k^e) + \\
    & \lambda_t \mathcal{L}_{BCE}((\mathcal{H}_s^*,\mathcal{H}_e^*), s_k^t)
\end{split}
\end{equation*}
For spatial grounding, smooth L1 loss, IoU loss and binary cross-entropy loss are used as follows,
\begin{equation*}
\begin{split}
    \mathcal{L}_k^{s} = & \lambda_l \mathcal{L}_{L_1}(\mathcal{B}^*,\mathcal{B}_k) + \lambda_i \mathcal{L}_{IoU}(\mathcal{B}^*,\mathcal{B}_k) + \\ 
    & \lambda_b \mathcal{L}_{BCE}(IoU(\mathcal{B}^*, \mathcal{B}_k),s_k^s)
\end{split}
\end{equation*}
The total training loss for training is $\mathcal{L}=\sum_{k=1}^{K}(\mathcal{L}_k^{t} +\mathcal{L}_k^{s})$.

\section{Experiments}

\noindent
{\bf Implementation.} Our CG-STVG is implemented using PyTorch. We use ResNet-101~\cite{resnet} as 2D backbone, VidSwin-tiny~\cite{vidswin} as 3D backbone, and RoBERTa-base~\cite{roberta} as text backbone. Following~\cite{STCAT,csdvl}, we utilize pre-trained MDETR~\cite{mdetr} to initialize the 2D backbone and text backbone. We use the Adam optimization algorithm~\cite{kingma2014adam} with a weight decay of $1e-4$ to end-to-end train our method. The initial learning rate for three backbones is set to $2e-5$ and $3e-4$ for the rest modules. We uniformly resize the video frames to a short side of $H$=$420$ and data augmentation methods such as random resizing and random cropping are applied to all training videos. The number of attention heads is set to $8$ and the hidden dimension of the encoder and decoder is $256$. The batch size is set to $16$ in HCSTVG-v1, $32$ in HCSTVG-v2 and $64$ in VidSTG dataset. The loss weight parameters $\lambda_s$, $\lambda_e$, $\lambda_t$, $\lambda_l$, $\lambda_i$, $\lambda_b$ are set to $10$, $10$, $1$, $5$, $3$, $1$, respectively. The number of decoding stages $K$ is set to $6$. We set the video frame length $N_v$ to $64$ and the text sequence length $N_t$ to $30$. The dimensions of the appearance feature, motion feature and text embedding $C_a$, $C_m$ and $C_t$ are $2048$, $768$, $768$. The temporal threshold $\theta^t$ and spatial threshold $\theta^s$ are set to $0.7$ and $0.8$, respectively.

\subsection{Datasets and Metrics.}

{\noindent \textbf{Datasets.}} Extensive experiments are conducted on three datasets, \ie, HCSTVG-v1~\cite{hcstvg}, HCSTVG-v2~\cite{hcstvg} and VidSTG~\cite{STGRN}. HCSTVG, focusing solely on humans in videos, is available in two versions: HCSTVG-v1 and HCSTVG-v2. Following~\cite{hcstvg,STCAT,TubeDETR}, we divide the HCSTVG-v1 into $4,500$ and $1,160$ video-sentence pairs for training and testing, respectively. HCSTVG-v2 further expands HCSTVG-v1, which includes $10,131$, $2,000$, and $4,413$ samples for training, validation, and testing, respectively. As the annotations for test set are not publicly available, we present the results based on validation set as existing methods~\cite{TubeDETR, csdvl}. VidSTG is another dataset constructed based on video relation dataset. Following~\cite{TubeDETR, STCAT, csdvl},  VidSTG is divided into training, validation, and test subsets with $80,684$, $8,956$, and $10,303$ distinct sentences, respectively, and $5,436$, $602$, and $732$ distinct videos, respectively.

\vspace{0.2em}
{\noindent \textbf{Metrics.}} Following~\cite{STVGBert,TubeDETR, STCAT}, we use m\_tIoU, m\_vIoU and vIoU@R as evaluation metrics. m\_tIoU measures temporal localization performance, while m\_vIoU and vIoU@R evaluate spatial localization. In specifc, m\_tIoU represents the average tIoU score over all testing sequences and tIoU is calculated as $\frac{|\mathcal{P}_i|}{|\mathcal{P}_u|}$, where $\mathcal{P}_i$ and $\mathcal{P}_u$ represent the intersection and union between the predicted segments and the ground-truth segments, respectively. Similarly, m\_vIoU represents the average vIoU score over all testing videos and vIoU is calculated as $\frac{1}{|\mathcal{P}_u|} \sum_{t \in \mathcal{P}_i} \text{IoU}(b^*_t, b_t)$, where $b^*_t$ and $b_t$ are the groundtruth bounding box and the predicted bounding box of the $t$-th frame. As for vIoU@R, it represents the ratio of samples with vIoU $\textgreater$ R in test subset.

\begin{table*}[ht]
	\centering
	\begin{minipage}{.5\textwidth}
		\centering
		\renewcommand{\arraystretch}{1.1}
		\scalebox{0.75}{
			\begin{tabular}{rcccc}
				\rowcolor{mygray} 
				\specialrule{1.5pt}{0pt}{0pt}
				Methods & m\_tIoU & m\_vIoU & vIoU@0.3 &  vIoU@0.5  \\ 
				\hline\hline
				STGVT \textcolor{lightgray}{\scriptsize{[TCSVT22]}}~\cite{hcstvg} & - &  18.2 & 26.8 & 9.5  \\
				STVGBert \textcolor{lightgray}{\scriptsize{[ICCV2021]}}~\cite{STVGBert} & - & 20.4 & 29.4 &  11.3  \\
				TubeDETR \textcolor{lightgray}{\scriptsize{[CVPR22]}}~\cite{TubeDETR} & 43.7 & 32.4 & 49.8 & 23.5 \\
				STCAT \textcolor{lightgray}{\scriptsize{[NeurIPS22]}}~\cite{STCAT} & 49.4 & 35.1 & 57.7 & 30.1 \\
				CSDVL \textcolor{lightgray}{\scriptsize{[CVPR23]}}~\cite{csdvl} & - & 36.9 & \textbf{62.2} & 34.8 \\ \hline
				Baseline & 50.4 & 36.5 & 58.6 & 32.3 \\ 
				CG-STVG & \textbf{52.8}\improve{~+2.4} & \textbf{38.4}\improve{~+1.9} & 61.5\improve{~+2.9} & \textbf{36.3}\improve{~+4.0} \\ \specialrule{1.5pt}{0pt}{0pt}
		\end{tabular}}
		\caption{Comparison with others on HCSTVG-v1 test set  (\%).}
		\label{tab:hcstvgv1}
	\end{minipage}%
	\hfill
	\begin{minipage}{.5\textwidth}
		\centering
		\renewcommand{\arraystretch}{1.1}
		\scalebox{0.75}{
			\begin{tabular}{rcccc}
				\specialrule{1.5pt}{0pt}{0pt}
				\rowcolor{mygray} 
				Methods & m\_tIoU & m\_vIoU & vIoU@0.3 &  vIoU@0.5  \\ \hline\hline
				PCC \textcolor{lightgray}{\scriptsize{[arxiv2021]}}~\cite{pcc} & - &  30.0 & - & -  \\ 
				2D-Tan \textcolor{lightgray}{\scriptsize{[arxiv2021]}}~\cite{2d-tan}  & - & 30.4 &  50.4 & 18.8  \\
				MMN  \textcolor{lightgray}{\scriptsize{[AAAI22]}}~\cite{mmn} & - & 30.3 & 49.0 & 25.6 \\
				TubeDETR \textcolor{lightgray}{\scriptsize{[NeurIPS22]}}~\cite{TubeDETR} & - & 36.4 & 58.8 & 30.6 \\
				CSDVL \textcolor{lightgray}{\scriptsize{[CVPR23]}}~\cite{csdvl} & 58.1 & 38.7 & \textbf{65.5} & 33.8 \\ \hline
				Baseline & 58.6 & 37.8 & 62.4 & 32.1 \\
				CG-STVG & \textbf{60.0}\improve{~+1.4} & \textbf{39.5}\improve{~+1.7} & 64.5\improve{~+2.1} & \textbf{36.3}\improve{~+4.2} \\
				\specialrule{1.5pt}{0pt}{0pt}
		\end{tabular}}
		\caption{Comparison with others on HCSTVG-v2 val. set  (\%).}
		\label{tab:hcstvgv2}
	\end{minipage}
\end{table*}

\begin{table*}[htb]
	\centering
	\renewcommand{\arraystretch}{1.06}
	\scalebox{0.87}{
		\begin{tabular}{rcccccccc}
			\specialrule{1.5pt}{0pt}{0pt}
			\rowcolor{mygray} 
			\cellcolor{mygray} & \multicolumn{4}{c}{ \cellcolor{mygray} Declarative Sentences} & \multicolumn{4}{c}{ \cellcolor{mygray}Interrogative Sentences} \\ 
			\rowcolor{mygray} 
			\multirow{-2}{*}{\cellcolor{mygray} Methods} & m\_tIoU & m\_vIoU & vIoU@0.3 &  vIoU@0.5  & m\_tIoU & m\_vIoU & vIoU@0.3 &  vIoU@0.5  \\
			\hline
			\hline
			STGRN \textcolor{lightgray}{\scriptsize{[CVPR20]}}~\cite{wheredoseit}  &  48.5 &  19.8 & 25.8 & 14.6 &  47.0 & 18.3 & 21.1 & 12.8 \\
			OMRN \textcolor{lightgray}{\scriptsize{[IJCAI20]}}~\cite{OAMBRN}  &  50.7 &  23.1 & 32.6 & 16.4 &  49.2 & 20.6 & 28.4 & 14.1 \\
			STGVT \textcolor{lightgray}{\scriptsize{[TCSVT22]}}~\cite{hcstvg} & - &  21.6 & 29.8 & 18.9 &  - & - & -  & - \\
			STVGBert \textcolor{lightgray}{\scriptsize{[ICCV21]}}~\cite{STVGBert}  & - &  24.0 & 30.9 & 18.4 & - & 22.5 & 26.0 & 16.0 \\
			TubeDETR \textcolor{lightgray}{\scriptsize{[CVPR22]}}~\cite{TubeDETR} & 48.1 &  30.4 & 42.5 & 28.2 & 46.9 & 25.7 & 35.7 & 23.2 \\
			STCAT \textcolor{lightgray}{\scriptsize{[NeurIPS22]}}~\cite{STCAT} & 50.8 & 33.1 & 46.2 & 32.6 & 49.7 & 28.2 & 39.2 & 26.6  \\
			CSDVL \textcolor{lightgray}{\scriptsize{[CVPR23]}}~\cite{csdvl} & - & 33.7 & 47.2 & 32.8 & - & 28.5 & 39.9 & 26.2  \\ \hline
			Baseline  & 49.7 & 32.4 & 45.0 & 31.4 & 48.8 & 27.7 & 38.7 & 25.6  \\
			CG-STVG & \textbf{51.4}\improve{~+1.7} & \textbf{34.0}\improve{~+1.6} & \textbf{47.7}\improve{~+2.7} & \textbf{33.1}\improve{~+1.7} &  \textbf{49.9}\improve{~+1.1} & \textbf{29.0}\improve{~+1.3} & \textbf{40.5}\improve{~+1.8} & \textbf{27.5}\improve{~+1.9} \\
			\specialrule{1.5pt}{0pt}{0pt}
	\end{tabular}}
	\caption{Comparison with existing state-of-the-art methods on VidSTG test set (\%).}
	\label{tab:vidstg}
	\vspace{-2mm}
\end{table*}

\begin{table}[h]
	\centering
	\renewcommand{\arraystretch}{1.1}
	\scalebox{0.92}{
		\begin{tabular}{ccccccc}
			\specialrule{1.5pt}{0pt}{0pt}
			\rowcolor{mygray} 
			ICG & ICR & m\_tIoU & m\_vIoU &  vIoU@0.3 &  vIoU@0.5 \\ \hline\hline
			- & - & 50.42 & 36.52 & 58.62 & 32.33 \\
			\checkmark & - & 51.07 & 37.42 & 59.48 & 32.93 \\
			\checkmark & T & 51.26 & 37.86 & 60.95 & 33.28 \\
			\checkmark & S & 52.80 & 38.04 & 60.90 & 35.40 \\
			\checkmark & S+T & \textbf{52.84} & \textbf{38.42} & \textbf{61.47} & \textbf{36.29} \\ \specialrule{1.5pt}{0pt}{0pt}
	\end{tabular}}
	\caption{Ablation study of ICG and ICR on HCSTVG-v1 test set of. ``T'' and ``S'' represent the temporal and spatial refinement.}
	\label{cmm}
	\vspace{-3mm}
\end{table}

\subsection{State-of-the-art Comparison}

{\noindent \textbf{HCSTVG-v1 and HCSTVG-v2.}} To validate the effectiveness of CG-STVG, we compare it with other state-of-the-arts on HCSTVG-v1 and HCSTVG-v2. 
Tab.~\ref{tab:hcstvgv1} shows the results on the HCSTVG-v1 test set, and our proposed method achieves state-of-the-art performance in 3 out of 4 metrics. Specifically, our method improves the $3.4$ absolute m\_tIoU score compared to STCAT~\cite{STCAT} and improves $1.5$ absolute m\_vIoU score compared to CSDVL~\cite{csdvl}. Compared to our baseline that does not use the proposed instance context by removing ICG and ICR modules, our method achieves improvements of $2.4$, $1.9$, $2.9$, and $4$ scores on the four metrics, respectively. 
On the validation set of the HCSTVG-v2, our method also achieves SOTA in 3 out of 4 metrics as shown in Tab.~\ref{tab:hcstvgv2}. CSDVL~\cite{csdvl} won the first place in the HCSTVG track of the 4-th Person in Context Challenge. Compared to the CSDVL, our approach outperforms it by $1.9$, $0.8$ and $2.5$ scores on m\_tIoU, m\_vIoU, and vIoU@0.5 metrics, respectively. The significant improvement in metric vIoU@0.5 across two datasets indicates that instance context excels at refining bounding boxes with an IoU under $0.5$.

\begin{table}
	\begin{minipage}{0.235\textwidth}
		\begin{subtable}{\textwidth}%
			\centering
			\renewcommand{\arraystretch}{1.1}
			\scalebox{0.68}{
				\begin{tabular}{cccc}
					\specialrule{1.5pt}{0pt}{0pt}
					\rowcolor{mygray} 
					$\theta^t$ & m\_tIoU &  m\_vIoU &  vIoU@0.5 \\ \hline\hline
					0.3 & 52.82 & 38.19 & 35.34 \\
					0.5 & 52.80 & 38.29 & 35.43 \\
					0.7 & \textbf{52.84} & \textbf{38.42} & \textbf{36.29} \\
					0.9 & 52.84 & 38.27 & 36.12 \\
					\specialrule{1.5pt}{0pt}{0pt}
			\end{tabular}}
			\caption{Ablation study for $\theta^{t}$.}
			\label{tab:ab2}
		\end{subtable}
	\end{minipage}
	\hfill
	\begin{minipage}{0.235\textwidth}
		\begin{subtable}{\textwidth}%
			\centering
			\renewcommand{\arraystretch}{1.1}
			\scalebox{0.68}{
				\begin{tabular}{ccccc}
					\specialrule{1.5pt}{0pt}{0pt}
					\rowcolor{mygray} 
					$\theta^s$ & m\_tIoU & m\_vIoU &  vIoU@0.5 \\ \hline\hline
					0.4 & 51.64 & 37.47 & 32.41 \\
					0.6 & 51.86 & 37.44 & 31.64 \\
					0.8 & \textbf{52.84} & \textbf{38.42} & \textbf{36.29} \\
					0.9 & 51.79 & 37.61 & 32.33 \\ 
					\specialrule{1.5pt}{0pt}{0pt}
			\end{tabular}}
			\caption{Ablation study for $\theta^{s}$.}
			\label{tab:ab1}
		\end{subtable}
	\end{minipage}
	\caption{Ablation of thresholds in ICR on HCSTVG-v1 test set.}
	\label{tab:threshold}
	\vspace{-3mm}
\end{table}

\begin{table}[!t]
	\centering
	\renewcommand{\arraystretch}{1.1}
	\scalebox{0.88}{
		\begin{tabular}{ccccc}
			\specialrule{1.5pt}{0pt}{0pt}
			\rowcolor{mygray}
			Usage of $s_{k}^{t}$ and $s_{k}^{s}$ &  m\_vIoU & vIoU@0.3 &  vIoU@0.5 \\ \hline\hline
			Two-level (ours) & \textbf{38.42} & \textbf{61.47} & \textbf{36.29} \\  
			one-level w/ ``$s_{k}^{t}+s_{k}^{s}$'' & 38.31 & 61.12 & 35.69 \\
			one-level w/ ``$s_{k}^{t}\times s_{k}^{s}$'' & 38.25 & 61.07 & 35.52 \\
			\specialrule{1.5pt}{0pt}{0pt}
	\end{tabular}}
	\caption{Ablation of usage of temporal and spatial confidence.}
	\label{fusion}
	\vspace{-3mm}
\end{table}

\begin{table}[!t]\small
	\centering
	\renewcommand{\arraystretch}{1.1}
	\scalebox{0.85}{
		\begin{tabular}{cccccc}
			\specialrule{1.5pt}{0pt}{0pt}
			\rowcolor{mygray}
			TDB & m\_tIoU &  m\_vIoU & vIoU@0.3 &  vIoU@0.5 \\ \hline\hline
			w/o Instance Context & \textbf{52.84} & \textbf{38.42} & \textbf{61.47} & \textbf{36.29} \\  
			w/ Instance Context & 52.61 & 38.01 & 61.03 & 35.78 \\
			\specialrule{1.5pt}{0pt}{0pt}
	\end{tabular}}
	\caption{Ablation on applying instance context to TDB.}
	\label{context_tdb}
	\vspace{-3mm}
\end{table}

\vspace{0.2em}
{\noindent \textbf{VidSTG Dataset.}} Besides HCSTVG-v1/-v2, we  compare CG-STVG with other methods on the challenging VidSTG dataset in Tab. \ref{tab:vidstg}. As shown, our method achieves the best results on all 8 metrics for both declarative sentences and interrogative sentences. With the proposed instance context, our method shows an improvement of $1.7$ m\_tIoU scores and $1.6$ m\_vIoU scores for declarative sentences and a gain of $1.1$ m\_tIoU scores and $1.3$ m\_vIoU scores for interrogative sentences over the baseline. The experimental results further evidence the effectiveness of our method, showing that instance context information helps ground the target.

\begin{figure*}[ht]
	\centering
	\includegraphics[width=\linewidth]{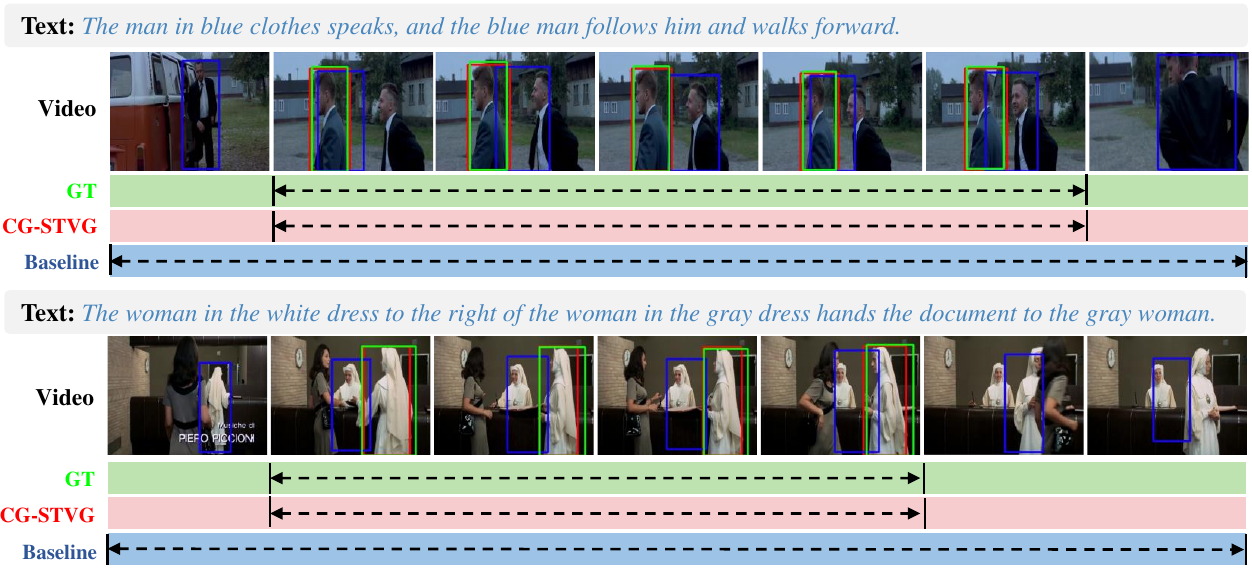}
	\caption{Qualitative results on HCSTVG-v1 test set. Our method (red) shows better localization than the baseline (blue).}
	\label{fig:qualitative}
	\vspace{-2mm}
\end{figure*}

\subsection{Ablation Study}

{\noindent \textbf{Impact of ICG and ICR.}} The key of CG-STVG lies in two simple yet effective modules, including ICG and ICR, for instance context learning. In order to verify their effectiveness, we conduct ablation experiments on HCSTVG-v1 in Tab.~\ref{cmm}. As in Tab.~\ref{cmm}, our baseline achieves a m\_vIoU score of $36.52$ without ICG and ICR module. After incorporating ICG for instance context, the m\_vIoU score is increased to $37.42$, demonstrating that the visual context extracted helps improve the grounding performance of the model. To enhance the quality of the instance context, we employ a spatial-temporal joint refinement mechanism in ICR module. When we applied temporal refinement alone, we observed that the m\_vIoU score improved by $0.44$. Applying the spatial refinement alone resulted in a slightly higher increase of $0.62$. However, when we employed both the spatial and temporal refinements simultaneously, the performance improvement is the most significant, with an increase of $1.0$, $37.42$ vs $38.42$. This demonstrates the synergistic effect of spatial and temporal refinements and underscores the effectiveness of our proposed spatial-temporal joint refinement mechanism in enhancing model performance.

\vspace{0.2em}
{\noindent \textbf{Impact of temporal and spatial thresholds in ICR.}}
In order to improve the quality of the instance visual context, we use the ICR module to filter the visual context from the ICG module. The ICR module refines the instance context through a two-level temporal-spatial joint refinement mechanism. Within this mechanism, there are two crucial parameters, temporal threshold $\theta^{t}$ and spatial threshold $\theta^{s}$, which are used as standards to filter the context. To investigate the influence of the temporal and spatial threshold on the performance of the model, we performed ablation experiments at different thresholds, as shown in Tab.~\ref{tab:threshold}. We can see that the model performs best when $\theta^{t}$ is $0.7$ and $\theta^{s}$ is $0.8$. 

\vspace{0.2em}
{\noindent \textbf{Impact of temporal and spatial confidence score usage.}} Temporal and spatial confidence scores $s_{k}^{t}$ and $s_{k}^{s}$ are crucial for instance context refinement. In this work, we adopt a two-level method to separately use $s_{k}^{t}$ and $s_{k}^{s}$ for refinement. To further study the impact of different methods for the usage of temporal and spatial confidence scores, we design two additional one-level methods for refinement: one is to add $s_{k}^{t}$ and $s_{k}^{s}$ and then apply a single fused confidence for refinement (one-level with ``$s_{k}^{t}+s_{k}^{s}$''), and the other is to multiple $s_{k}^{t}$ and $s_{k}^{s}$ for refinement (one-level with ``$s_{k}^{t}\times s_{k}^{s}$''). We show the architectures of these two variants and comparison with our strategy in \textbf{supplementary material} due to limited space. We conduct experiments on HCSTVG-v1 in Tab.~\ref{fusion}, and we can see that our two-level method to separately use $s_{k}^{t}$ and $s_{k}^{s}$ achieves better performance.

\vspace{0.2em}
{\noindent \textbf{Impact of applying instance context to TDB.}} From the Tab.~\ref{cmm}, it can be seen that as the spatial grounding improves with the help of context, the temporal grounding is also improving, $50.42$ vs $52.84$. To explore the impact of applying instance context to the TDB on model performance, we conduct ablation study as shown in Tab.~\ref{context_tdb}. There is a slight drop in model performance after employing context to the TDB. We believe the temporal branch is mainly used to determine the boundaries of events, and the context from the spatial branch has a gap with the temporal branch. Directly using context in temporal branch may cause boundary blur.

\subsection{Qualitative Analysis}

We present qualitative results in Fig.~\ref{fig:qualitative}. Compared to our baseline, CG-STVG could accurately locate the target temporally and spatially with instance context. More results can be seen in the \textbf{supplementary material}.

\section{Conclusion}
In this work, we introduce CG-STVG, a novel and simple method for improving STVG via exploiting instance visual context from video to guide target localization. The strength of CG-STVG comes from two key modules, including ICG that mines coarse visual context, and ICR that refines this context using time and space information. The experimental results on three challenging benchmarks further demonstrate the effectiveness of our method.

{
\small
\bibliographystyle{ieeenat_fullname}
\bibliography{main}
}

\maketitlesupplementary

\noindent
In this supplementary material, we present more details and analysis as well as results of our work, as follows,
\begin{itemize}
	\setlength{\itemsep}{2pt}
	\setlength{\parsep}{2pt}
	\setlength{\parskip}{2pt}
	
	\item[] \textbf{A. Detailed Architectures of Modules} \\ We display the detailed architectures for $\mathtt{SAEncoder}(\cdot)$, $\mathtt{SA}(\cdot)$ and $\mathtt{CA}(\cdot)$ in the main text. In addition, we present the architectures for different usage of temporal and spatial confidence scores.
	
	\item[] \textbf{B. Additional Ablation on Motion Information} \\ We conduct an extra experiment to ablate motion information in our approach.
	
	\item[] \textbf{C. More Visualization Analysis on Attention Maps} \\ We include more visualization analysis on the attention maps to show the effectiveness of instance context in improving target-awareness for localization.

 \item[] \textbf{D. More Qualitative Results} \\ We demonstrate more qualitative results of our method for grounding the target object.
	
\end{itemize}

\section{Detailed Architectures of Modules}
\label{details}

\subsection{Architecture of Self-Attention Encoder}
\begin{figure}[h]
	\centering
	\includegraphics[ width=0.85\linewidth]{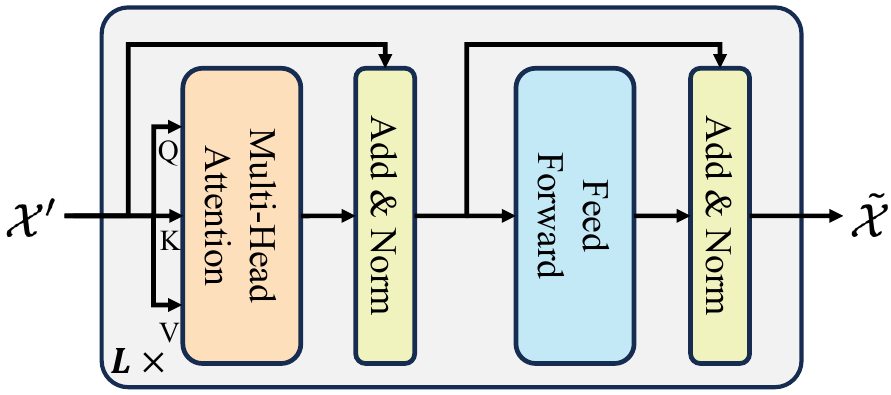}
	\caption{Detailed architecture of $\mathtt{SAEncoder}(\cdot)$.}
	\label{fig:saencoder}
\end{figure}

The self-attention encoder module, \ie, $\mathtt{SAEncoder}(\cdot)$, is to enhance multimodal feature $\mathcal{X}'$ and output $\mathcal{\tilde{X}}$, which is composed of $L$ ($L$=6) standard self-attention blocks, as depicted in Fig.~\ref{fig:saencoder}. 

\subsection{Architectures of Attention Blocks in Decoder}
\begin{figure}[!t]
	\centering
	\includegraphics[width=1.0\linewidth]{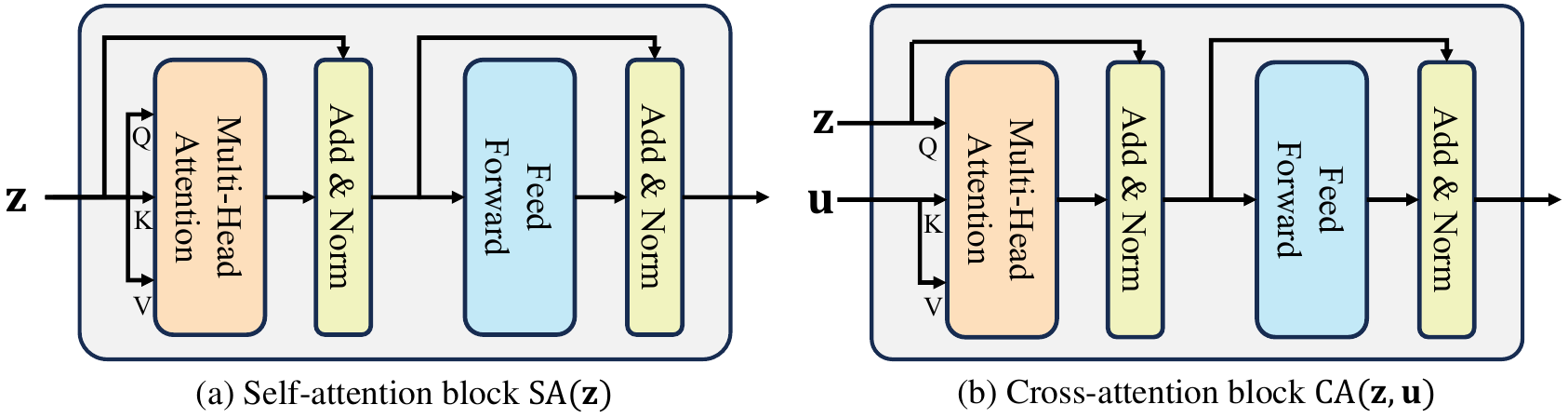}
	\caption{The architectures $\mathtt{SA}(\textbf{z})$ and $\mathtt{CA}(\textbf{z}, \textbf{u})$ in our model.}
	\label{fig:modules}
\end{figure}

In our context-guided decoder, we employ attention blocks, including the self-attention block, \ie, $\mathtt{SA}(\textbf{z})$ and the cross-attention block, \ie, $\mathtt{CA}(\textbf{z}, \textbf{u})$. There architectures are shown in Fig.~\ref{fig:modules}.

\begin{figure}[!h]
	\centering
	\includegraphics[ width=0.9\linewidth]{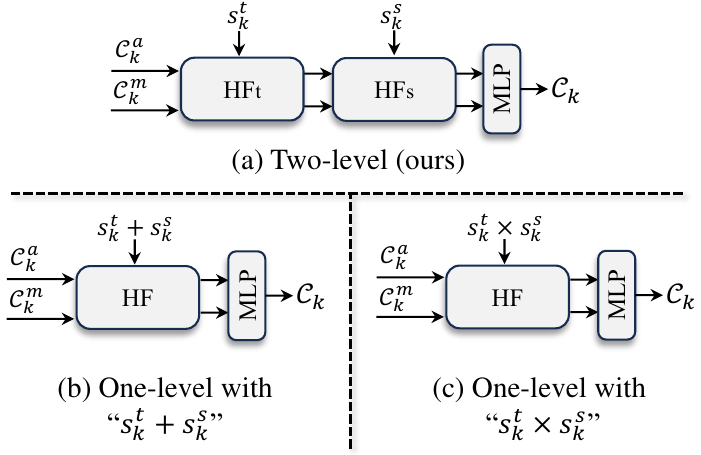}
	\caption{Architectures for different usage of spatial and temporal confidence scores. Image (a) is the proposed two-level architecture, and image (b) and (c) are two one-level variants with ``addition'' and ``multiplication'', respectively. }
	\label{fig:fusion}
\end{figure}

\begin{table}[!th]
\centering
\renewcommand{\arraystretch}{1.3}
\scalebox{0.78}{
\begin{tabular}{cccccc}
\rowcolor{mygray} 
\specialrule{1.5pt}{0pt}{0pt}
\multicolumn{2}{c}{\cellcolor{mygray}Motion} & \cellcolor{mygray} & \cellcolor{mygray} & \cellcolor{mygray} & \cellcolor{mygray}  \\
\rowcolor{mygray}
Feature & Context & \multirow{-2}{*}{\cellcolor{mygray}m\_tIoU} & \multirow{-2}{*}{\cellcolor{mygray}m\_vIoU} & \multirow{-2}{*}{\cellcolor{mygray}vIoU@0.3} &  \multirow{-2}{*}{\cellcolor{mygray}vIoU@0.5} \\ \hline\hline
- & - & 49.28 & 35.81 & 58.36 & 29.91 \\  
\checkmark & - & 51.56 & 37.62 & 59.57 & 32.16 \\  
\checkmark & \checkmark  & \textbf{52.84} & \textbf{38.42} & \textbf{61.47} & \textbf{36.29} \\
\specialrule{1.5pt}{0pt}{0pt}
\end{tabular}}
\caption{Ablation of motion information on HCSTVG-v1 (\%).}
\label{motion}
\end{table}

\subsection{Architectures for Different Usage of Temporal and Spatial Confidence Scores}
In the ablation, we compare our two-level strategy with two additional one-level strategiess for exploiting the temporal and spatial confidence scores. The structures of these three mechanisms are compared and illustrated in Fig.~\ref{fig:fusion}.

\vspace{-2pt}
\section{Ablation on Motion Information}
\label{ablation}
Motion information is complementary to appearance cues and can benefit STVG. Specifically, it provides a few extra advantages: (1) Motion features contain the movement details of the target object, which are crucial for the STVG task; (2) Motion cues can provide useful temporal information to some extent even when the appearance partially invisible; (3) Motion features can better comprehend the spatial relationships between objects in the video, such as distance and relative position. Thus inspired, we utilize both appearance and motion features in our CG-STVG, as in many other STVG methods. To study the impact of motion information in our CG-STVG, we provide additional ablation results in Tab.~\ref{motion}. As shown in Tab.~\ref{motion}, with the help of the motion features $\mathcal{\tilde{X}}_{m}$ in multimodal feature $\mathcal{\tilde{X}}$, the m\_vIoU increases by $1.81$, achieving $37.62$, which shows that motion features can provide the necessary action information for STVG. After integrating motion context extracted from motion features, the m\_vIoU score has improved to $38.42$, demonstrating the effectiveness of motion context.

\begin{figure}[!h]
	\centering
	\includegraphics[ width=0.95\linewidth]{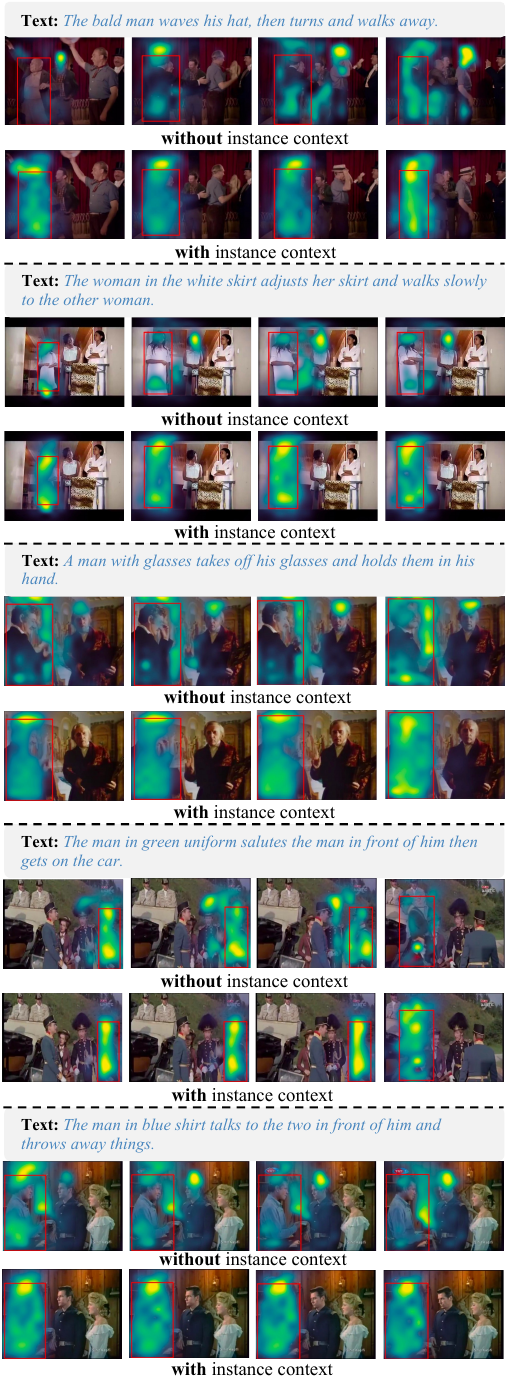}
	\caption{Attention maps for spatial queries in video frames in the spatial-decoding block \emph{without} and \emph{with} our proposed instance context. The red boxes indicate the foreground object to localize.}
	\label{fig:att}
\end{figure}

\section{More Visualization on Attention Maps}
\label{visualize}
In order to analyze the role of instance context, we compare the attention maps of spatial queries in the spatial-decoding block (SDB), with and without using instance context, as in Fig.~\ref{fig:att}. From Fig.~\ref{fig:att}, we can clearly see that, the queries, without being enhanced by the instance context, are unable to focus on the foreground object across different frames. However, when employing instance context, the queries are significantly enhanced by gaining more target-awareness knowledge to focus on the foreground regions, which benefits accurate localization of the target object and thus improves the STVG performance.

\section{More Qualitative Results}
\begin{figure*}[t]
	\centering
	\includegraphics[trim=0 40 500 0, clip, width=1.0\linewidth]{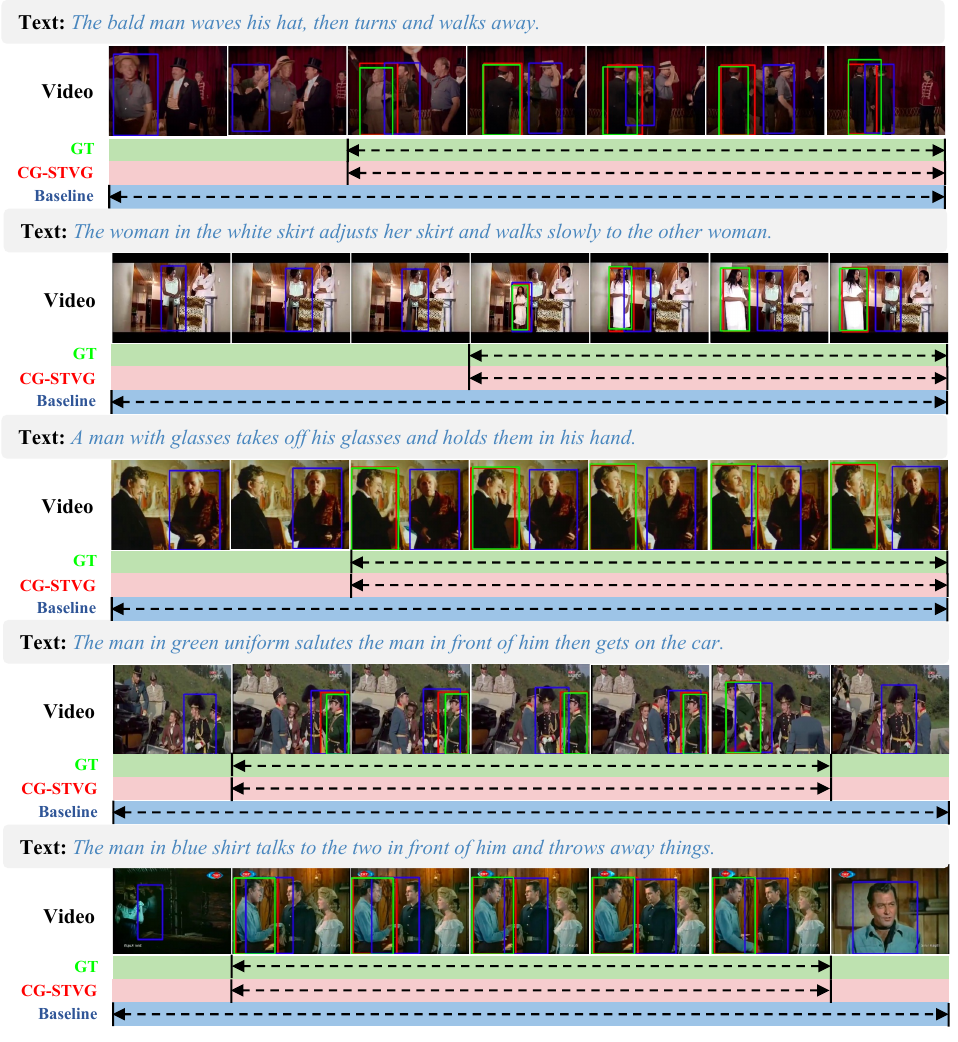}
	\caption{Qualitative results of our method (red), our baseline method (blue) and ground truth (green). In all examples, our method shows better localization than the baseline, which proves the importance of instance context.}
	\label{fig:res}
\end{figure*}

To further validate the effectiveness of our method (with instance context), we provide additional examples of grounding results compared to the baseline method (without instance context) on the HCSTVG dataset in Fig.~\ref{fig:res}.  From the shown visualizations, the baseline model struggles to locate the target object accurately within the video frames. However, when employing the mined instance visual context, our method is able to localize the target object with better temporal and spatial accuracy. 

In detail, for the second example, the key words in the text ``The woman in the white skirt adjusts her skirt and walks slowly to the other woman'' are ``white" and ``walks''. However, since there are three women wearing white in the video, the information of ``white" may not be discriminative and useful. Therefore, we can merely rely on other information such as ``walks" and only the fourth frame of the video contains information about ``walks". As a result, our method and the baseline method can both accurately locate the target in the fourth frame. However, since the baseline method does not have context guidance, there is no reliable information to use in the remaining frames, leading to errors in time and space localization. Our method, on the other hand, accurately locates the target by using the mined instance context from the fourth frame to assist in locating the target in the remaining frames. The same comparison can also be observed in the fourth example. These examples further demonstrate the importance of instance context in guiding target localization.

\end{document}